\documentclass[sigconf]{acmart}
\AtBeginDocument{%
  }

\setcopyright{acmlicensed}
\copyrightyear{2026}
\acmYear{2026}
\acmDOI{XXXXXXX.XXXXXXX}
\acmConference[KDD '26]{32nd ACM SIGKDD Conference on Knowledge Discovery and Data Mining}{August 09--13,
  2026}{Jeju, Korea}
\acmISBN{978-1-4503-XXXX-X/2018/06}

\usepackage{times}
\usepackage{latexsym}

\usepackage[T1]{fontenc}

\usepackage[utf8]{inputenc}

\usepackage{microtype}

\usepackage{graphicx}

\usepackage{multirow}
\usepackage{amsmath}
\usepackage{enumitem}
\usepackage{amsmath}
\usepackage{subcaption}
\usepackage{float}
\usepackage{booktabs}

\usepackage{algorithm}
\usepackage{algorithmic}

\usepackage{hyperref}
\usepackage{makecell}
\usepackage{arydshln}

\newcommand{\first}{\colorbox[HTML]{FDD835}}
\newcommand{\second}{\colorbox[HTML]{FFEE58}}
\newcommand{\third}{\colorbox[HTML]{FFF9C4}}

\usepackage{colortbl}
\usepackage{xcolor}
\usepackage[table]{xcolor}
\usepackage{multirow}
\usepackage{booktabs}
\usepackage{array}

\newcommand{\dataset}{{\fontfamily{cmss}\selectfont SUDO}}

\begin{document}

\title{Comparing Without Saying: A Dataset and Benchmark for Implicit Comparative Opinion Mining from Same-User Reviews}

\author{Thanh-Lam T. Nguyen}
\email{22024516@vnu.edu.vn}
\affiliation{
  \institution{c and Technology}
  \city{Hanoi}
  \country{Viet Nam}
}

\author{Ngoc-Quang Le}
\email{22024510@vnu.edu.vn}
\affiliation{
  \institution{VNU University of Engineering and Technology}
  \city{Hanoi}
  \country{Viet Nam}
}

\author{Quoc-Trung Phu}
\email{21020096@vnu.edu.vn}
\affiliation{
  \institution{VNU University of Engineering and Technology}
  \city{Hanoi}
  \country{Viet Nam}
}

\author{Thi-Phuong Le}
\email{tp.le.2023@phdcs.smu.edu.sg}
\affiliation{%
  \institution{Singapore Management University}
  \country{Singapore}
}

\author{Ngoc-Huyen Pham}
\email{23020541@vnu.edu.vn}
\affiliation{
  \institution{VNU University of Engineering and Technology}
  \city{Hanoi}
  \country{Viet Nam}
}

\author{Phuong-Nguyen Nguyen}
\email{24022702@vnu.edu.vn}
\affiliation{
  \institution{VNU University of Engineering and Technology}
  \city{Hanoi}
  \country{Viet Nam}
}

\author{Hoang-Quynh Le}
\email{lhquynh@vnu.edu.vn}
\affiliation{
  \institution{VNU University of Engineering and Technology}
  \city{Hanoi}
  \country{Viet Nam}
}

\renewcommand{\shortauthors}{Nguyen et al.}

\begin{abstract}
    Existing studies on comparative opinion mining have mainly focused on explicit comparative expressions, which are uncommon in real-world reviews. This leaves implicit comparisons -- where users express preferences across separate reviews -- largely underexplored. We introduce \dataset{}, a novel dataset for implicit comparative opinion mining from same-user reviews, allowing reliable inference of user preferences even without explicit comparative cues. \dataset{} comprises $4,150$ annotated review pairs ($15,191$ sentences) with a bi-level structure capturing aspect-level mentions and review-level preferences. We benchmark this task using 
    two baseline architectures: Traditional Machine Learning- and Language Model-based baselines. 
    Experimental results show that while 
    the latter outperforms the former, overall performance remains moderate, revealing the inherent difficulty of the task and establishing \dataset{} as a challenging and valuable benchmark for future research.
\end{abstract}




\maketitle

\section{Introduction}
\begin{figure}[t!]
  \centering
  \includegraphics[width=1.0\linewidth]{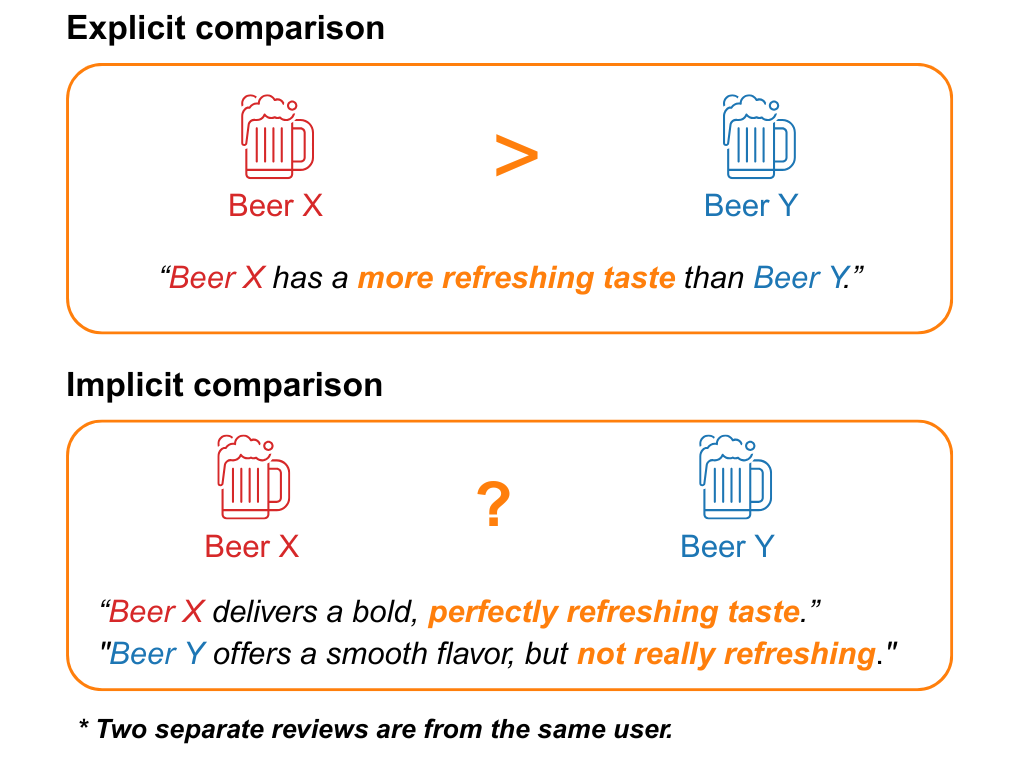}
  \caption{Illustration of explicit vs. implicit comparative opinions in user reviews.}
  \raggedright
  \small{\textbf{Explicit:} Direct comparison using terms like `\textit{more refreshing than}'. 
  \textbf{Implicit:} Two separate reviews from the same user. Preference is implied -- harder to detect.}
  \label{fig:implicit_comparison}
\end{figure}

In today's market, where countless brands and options are available, users often struggle to make informed decisions. Traditional reviews primarily capture individual experiences, but rarely offer direct comparisons between alternatives. 
Although consumers may have experienced multiple products, they rarely provide explicit comparisons in their reviews. 
Comparative opinion mining addresses this gap by analyzing and comparing different options. It helps users discern key differences and make more informed choices, particularly when evaluating similar products ~\textcolor{brown}{\cite{varathan2017comparative}}.
As illustrated in Figure~\textcolor{brown}{\ref{fig:implicit_comparison}}, user opinions can be expressed through explicit or implicit comparisons. While explicit comparisons clearly state a preference between products, implicit ones are often spread across separate reviews, requiring contextual inference for interpreting user preferences. Such reviews may contain valuable insights, but without structured comparisons, it is difficult for users to determine which aspects matter most or which product performs better overall. As a result, users often struggle to extract relevant insights from an overwhelming volume of fragmented reviews. 
Given two reviews written by the same user for different products, the task of implicit comparative opinion mining is to infer the preferred product for each defined aspect. We prioritize the same-user setting to mitigate noise arising from individual behavioral variances.
Consider a strict user who gives low ratings and writes brief summaries, versus a lenient user known for high ratings and detailed reviews. Paradoxically, a rating of \textit{``good''} from the former often implies higher quality than \textit{``amazing''} from the latter. By isolating comparisons to the same user, we remove inconsistencies and capture preferences against a baseline.

Although several comparative opinion mining datasets have been proposed in recent years, most of them primarily focus on explicit comparisons, where comparative statements are explicitly expressed in a single review. 
As a result, the subtler and more realistic case of implicit comparisons remains underexplored. Some recent efforts have attempted to address implicit comparisons by pairing reviews of different products but these datasets typically involve reviews 
written by different users. Therefore, the paired reviews often differ significantly in writing style, personal preferences, and contextual structure.
These inconsistencies make it difficult to isolate the true comparative intent and understand implicit preferences clearly.


To address this gap, we introduce \textbf{\dataset{} }(\textbf{S}ame \textbf{U}ser -- \textbf{D}ifferent \textbf{O}bjects/products), a dataset that captures implicit comparative preferences of the same user across different products. \dataset{} consists of review pairs written by the same user, allowing inference of subtle preferences without considering user-specific differences. 
It features a hierarchical bi-level annotation scheme in which sentences are annotated with aspect-level labels, while comparative opinions are annotated at the review level.
To our knowledge, \dataset{} is the first dataset to systematically capture implicit product comparisons from the perspective of the same user.
As baselines, we present two architectures -- Traditional Machine Learning- and Language Model-based baselines -- to address the problem of comparing two reviews across aspects. 

The dataset, models, and source code will be made publicly available upon the formal acceptance and publication of this work.


\section{Related Work}
Comparative opinion mining~\textcolor{brown}{\cite{jindal2006mining}} was pioneered by extracting comparative sentences from user-generated content, introducing two key subtasks: sentence identification and relation extraction, paving the way for later research.
Opinion orientation in comparative sentences~\textcolor{brown}{\cite{ganapathibhotla2008mining}} was addressed by a rule-based method that identified which entity was preferred by the author using domain knowledge and contextual cues. However, it did not account for complex structures or inter-sentence comparisons commonly found in user reviews.

The Comparative Opinion Quintuple Extraction (COQE) task and dataset ~\textcolor{brown}{\cite{liu2021comparative}} -- which extend traditional comparative mining by extracting all comparative opinion quintuples (Subject, Object, Comparative Aspect, Comparative Opinion, Comparative Preference) -- were presented. A multistage BERT-based approach significantly outperformed baseline systems in three datasets. Nevertheless, inter-sentence comparisons were not the focus of this work. The CoCoTrip dataset ~\textcolor{brown}{\cite{hayate2022comparative}} -- which utilizes summarization to highlight contrastive and common points between hotels based on multiple user reviews -- is tangentially related. Although not focused on opinion mining, it shares the goal of extracting differences between entities. However, it is relatively small (48 pairs) and relies on human-written summaries, limiting natural linguistic variation. The VCOM corpus and benchmark model in the Vietnamese smartphone domain ~\textcolor{brown}{\cite{can2025toward}} represent a notable contribution to comparative opinion mining. The dataset focuses on explicit comparative structures extracted from online product reviews, comprising over 2,400 annotated comparative tuples. Its reliance on overt comparative language limits the usefulness in real-world scenarios dominated by implicit preferences, making it less suitable for capturing subtle comparative intents.

Despite significant progress, most existing studies remain constrained by their reliance on explicit comparative sentences and limited ability to handle arbitrary product pairs. Our dataset and baselines address these challenges by uncovering implicit comparative opinions embedded in same-user review pairs. We also provide a detailed comparison between the \dataset{} dataset and other related corpora to highlight its unique characteristics and applicability in Section~\textcolor{brown}{\ref{sec:datacompare}} and Table~\textcolor{brown}{\ref{tab:comparisons_btw_datasets}}.

\section{Dataset Construction}

\subsection{Annotation Pipeline}
\label{subsec:annotation-pipeline}
\begin{figure}[!h]
  \includegraphics[width=\columnwidth]{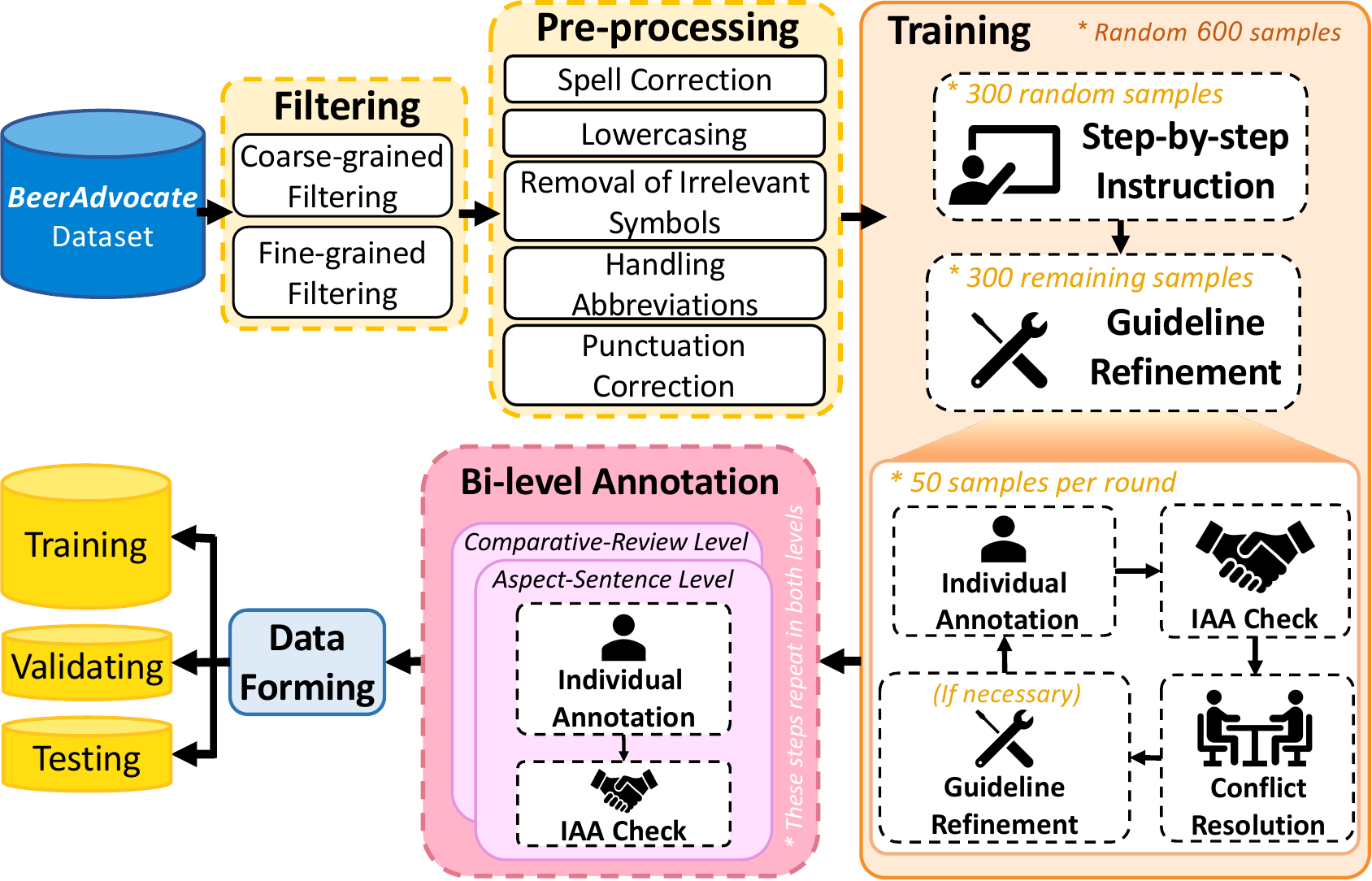}
  \caption{Data annotation pipeline.}
  \label{fig:annotation_pipeline}
\end{figure}

Our \dataset{} dataset is built upon the BeerAdvocate corpus~\textcolor{brown}{\cite{mcauley2013from}}, whose reviews naturally cover four sensory aspects: \texttt{appearance}, \texttt{aroma}, \texttt{palate}, and \texttt{taste}. This dataset consists of user-generated beer reviews in structured tabular form and contains approximately 1.5 million reviews collected over a decade. We chose the beer domain because its reviews are rich in aspect-level descriptions and highly detailed, making them well-suited for analyzing implicit comparative opinions.
Moreover, the dataset’s large-scale, user-identified reviews enable the construction of same-user pairs, which is a key requirement for our task design.

Since real-world reviews often intertwine these aspects with personal impressions, we designed a two-step annotation process (see Figure~\textcolor{brown}{\ref{fig:annotation_pipeline}}). In the first step, each sentence is examined to determine whether it explicitly refers to any of the defined aspects. In the second step, only the aspect-specific sentences are grouped into review pairs written by the same user. Our pairing strategy involved selecting users with $5$ reviews and generating all possible round-robin combinations for these users, forming $10$ pairs per user in the final dataset. These pairs are then labeled with comparative information.

\subsubsection{Data Filtering and Preprocessing}
\paragraph{Data Filtering}  
To construct the \dataset{} dataset, we performed a two-stage filtering.

\textbf{Coarse-Grained Filtering:}  
We first filtered out users with insufficient review activity. Specifically, we retained only users who have written at least five reviews, providing a sufficient basis for constructing multiple review pairs in subsequent steps. This threshold ensures that pairwise comparisons can be made while ensuring consistent writing style and evaluation criteria for each user.

\textbf{Fine-Grained Filtering:}
We manually inspected all reviews to remove those lacking descriptive content. Specifically, we excluded vague or generic opinions (e.g., \textit{``good beer''}, \textit{``pretty decent''}) that did not elaborate on the four sensory aspects. This quality-assurance step posed two main challenges. First, it was time-consuming and labor-intensive, requiring consistent manual filtering across thousands of reviews, essential for reliability but limiting scalability. Second, many reviews conveyed aspect-related content implicitly (e.g., \textit{``fizzes out quickly in the glass''} implies appearance; \textit{``left a sticky feeling after the last sip''} implies palate), demanding annotators to interpret subtle sensory cues and apply domain knowledge.

\paragraph{Preprocessing}  
We applied standard text preprocessing steps to ensure review quality and overall data consistency. These steps included lowercasing, correcting common misspellings, removing non-text characters, expanding common slang and abbreviations, and removing redundant or misplaced punctuation.

\subsubsection{Data Annotation}
\paragraph{Training Annotators} 
The annotation was carried out by a team of four native-speaking students who were carefully selected and trained for this task. The process was supervised by a Ph.D. in Natural Language Processing.
The team included one Ph.D. student in Computational Linguistics, one graduate student in Information Technology, and two undergraduate students majoring in Information Systems. 
Annotators underwent guided practice on $300$ samples to understand the dataset’s structure and linguistic nuances.

An annotation guideline with detailed instructions, developed by linguistic experts, provided the foundation and was iteratively refined during the process. To ensure consistency, annotation began with an initial calibration phase on the first $300$ samples, where at least two annotators labeled each item. After every batch of $50$ samples, Inter-Annotator Agreement (IAA)~\cite{steinert2017collaborative} was evaluated. Annotators then discussed disagreements to further refine the guideline. After achieving stable agreement, the remaining samples were annotated by two annotators per item following the finalized guideline.

The final IAA, calculated over the entire dataset, was $0.97$ at the sentence–aspect level and $0.85$ at the review–comparative level, indicating reliable and consistent annotations.

\paragraph{Bi-level Data Annotation} 
The dataset comprises pairs of beer reviews written by the same users. Its main purpose is to determine whether one review expresses a more favorable, less favorable, similar, or incomparable opinion relative to the other.

Each review includes ratings for four aspects $\mathcal{A} = \{\texttt{appearance}, \allowbreak \texttt{aroma}, \texttt{palate}, \texttt{taste}\}$.
The dataset was annotated at two levels:
\begin{enumerate}[noitemsep,topsep=0pt]
    \item \textbf{Sentence-Aspect Level}: Each sentence is labeled as \texttt{1} (Mentioned) if it refers to a specific aspect, or \texttt{0} (Not Mentioned) otherwise.
    \item \textbf{Review-Comparative Level}: Each review pair is labeled per aspect as follows: \texttt{-1} (worse -- the first review is rated lower than the second for that aspect), \texttt{0} (similar or non-comparable), \texttt{1} (better -- the first review is rated higher than the second), or \texttt{Null} (the aspect is not mentioned in one or both reviews).

\end{enumerate}
For pairs where the aspect is not mentioned in sentence-aspect level, the comparative label is automatically set to null label. These cases are excluded from Phase 2 training and evaluation because they carry no comparative information, so their high frequency does not introduce bias into model learning.

After the training, the remaining data were divided into subsets, each annotated consistently following the same two-level procedure. First, all annotators independently labeled the same subset. Second, IAA was evaluated. If it reached at least $0.8$, the process for that subset was complete. Otherwise, annotators collaboratively reviewed the subset to resolve disagreements. In such cases, the final decision typically chose the lower score to maintain conservative labeling.

\paragraph{Data Forming} 
We split the dataset at the user level, assigning $80\%$ of users to the training set, $10\%$ to the validation set, and the remaining $10\%$ to the test set. This prevents information leakage and ensures generalization to unseen users. 

\begin{table*}[t!]
\centering
\small
\renewcommand{\arraystretch}{1.4}

\caption{The comparison between \dataset{} dataset and some prior datasets on comparative opinion mining.}

\resizebox{0.9\linewidth}{!}{%
\begin{tabular}{@{}p{3cm} p{1.8cm} p{1.8cm} p{1.8cm} p{1.8cm} p{1.8cm} p{1.8cm}@{}}
\toprule
\textbf{} & \textbf{\dataset{}\newline(our data)} & \textbf{VCOM (2015)} & \textbf{CocoTrip (2022)} & \textbf{Bach et al. (2015)} & \textbf{Jindal \& Liu (2006)} & \textbf{Kessler \& Kuhn (2014)} \\
\midrule
Implicit\slash Explicit & Implicit & Explicit & Implicit & Explicit & Explicit & Explicit \\
\midrule
Aspect Annotation Scheme & Classification & Extraction & N/A & Extraction & Extraction & Extraction \\
\midrule
Comparison Annotation Scheme & Classification & Extraction\slash Classification & Summarization & Extraction\slash Classification & Extraction\slash Classification & Extraction \\
\midrule
Intra-\slash Inter-sentence$^*$ & Inter & Intra\slash Inter & Inter & Intra & Intra & Intra \\
\midrule
Domain & Beer & Smartphone & Hotels & Electronic products & General products & Camera \\
\midrule
\#Reviews & 2075 & 120 & 768 & N/A & N/A & 564 \\
\midrule
\#Sentences & 15191 & 9174 & 3000 & 4000 & 3248 & 11232 \\
\midrule
\#Aspects & 4 & 2109 & 6 & 2942 & 348 & 1407 \\
\midrule
\#Comparison Labels & 5 & 8 & N/A & 3 & 4 & 4 \\
\bottomrule
\multicolumn{7}{r}{$^*$Intra-sentence: labeling within a single sentence; Inter-sentence: captures information spanning multiple sentences.}
\end{tabular}
}

\label{tab:comparisons_btw_datasets}
\end{table*}

\subsection{Comparisons With Other Datasets}
\label{sec:datacompare}
Table~\textcolor{brown}{\ref{tab:comparisons_btw_datasets}} compares the \dataset{} dataset with typical datasets in the field. The primary goal of \dataset{} is to focus on implicit expressions, often overlooked in existing resources.
Among the compared datasets, CoCoTrip~\textcolor{brown}{\cite{hayate2022comparative}} partially captures implicit information via contrastive summarization of 48 hotel pairs, but it focuses on summary generation rather than aspect-level implicit comparison and lacks the structural detail for fine-grained opinion analysis.
\dataset{} is distinguished not only by its larger scale but also by its bi-level manual annotation covering both aspect and comparative sentiment.
Both annotation levels are formulated as classification tasks, which provide more abstract and transferable representations compared to extraction-based schemes that rely on surface linguistic cues.
\dataset{} also includes inter-sentence comparisons to capture patterns where comparisons are expressed across multiple sentences. These characteristics enhance the reliability and real-world applicability of \dataset{}, and offer additional advantages for robust model training, improved generalization, and more comprehensive downstream comparative opinion analysis.

\vspace{-1em}

\section{Data Statistics and Analysis} 
\label{sec:data_statistics_analysis}
\subsection{Label Distribution}
\begin{figure}[h]
\centering
  \includegraphics[width=0.96\columnwidth]{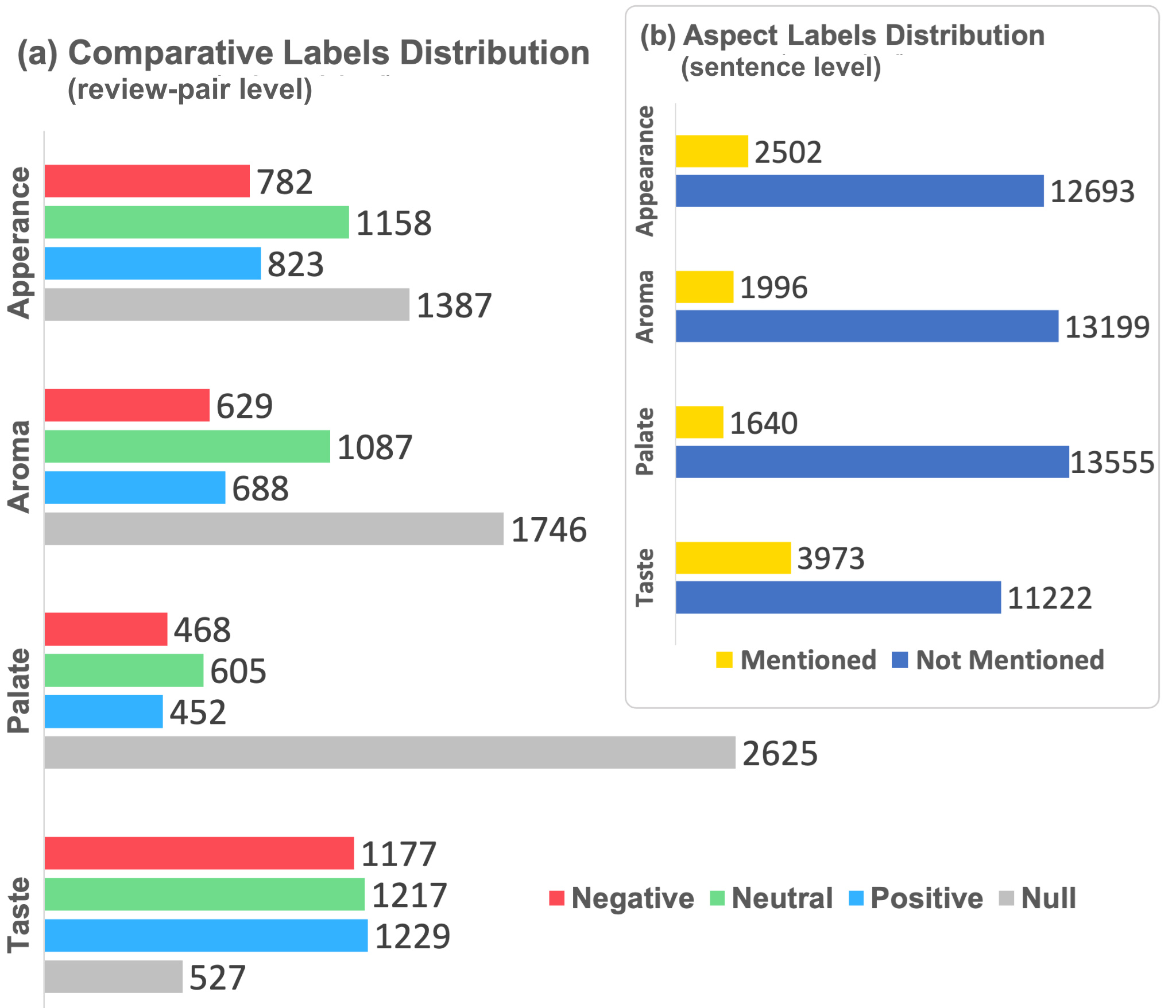}
  \caption{Distribution of aspect and comparative labels.}
  \label{fig:labels_distribution}
\end{figure}

In Figure~\textcolor{brown}{\ref{fig:labels_distribution}} (b), the distribution of aspect mentions at the sentence level shows a clear imbalance. Sentences that do not mention a given aspect far outnumber those that do, highlighting the difficulty of accurately detecting aspect-related sentences and emphasizing the need for robust aspect identification methods.
Among all aspects, \texttt{taste} is the most frequently mentioned, showing its importance in reviewers’ beer descriptions and reflecting the tendency to focus evaluations on sensory experience.

As shown in Figure~\textcolor{brown}{\ref{fig:labels_distribution}} (a), a large portion of review pairs are labeled \texttt{Null}, indicating that the corresponding aspect is not mentioned in one or both reviews. The prevalence of \texttt{Null} varies by aspect: \texttt{taste} shows the fewest \texttt{Null} instances, highlighting its central role in beer evaluation, whereas \texttt{palate} exhibits the largest proportion of \texttt{Null} cases, indicating it is infrequently discussed.
Considering non-null pairs, the sentiment labels are slightly skewed toward \texttt{Neutral}, suggesting that when both reviews mention an aspect, they often express similar sentiments. \texttt{Positive} and \texttt{Negative} labels are both present and roughly balanced in magnitude. This balance is expected, as whether a comparison is labeled \texttt{positive} or \texttt{negative} depends on the arbitrary ordering of the two reviews.
Although \texttt{appearance} is mentioned less often overall, the distribution of its pairs across sentiment classes is relatively balanced, suggesting that when reviewers comment on appearance, they often provide varied comparative judgments.
Overall, the high proportion of \texttt{Null} and \texttt{Neutral} pairs poses challenges for supervised learning. Specifically, this imbalance creates a tendency for models to over-predict the majority classes, effectively ignoring the minority cases to maintain high accuracy. This bias makes it difficult for the model to learn the subtle distinctions required to capture actual comparative preferences.

\subsection{Cross-Aspects Relationship Analysis}
\label{subsec:aspects_interdependencies_analysis}

\begin{table}[H]
    \centering
    
    \caption{Chi-Square inspection on the dependency between aspects.}
    
    \resizebox{\linewidth}{!}{%
        \begin{tabular}{llll}
            \hline
                \textbf{Aspect 1} & \textbf{Aspect 2} & \textbf{Chi-Sq value} & \textbf{p-value} \\ \hline
                appearance & aroma  & 281.781963  & \(9.20 \times 10^{-60}\) \\
                appearance & palate & 279.893618  & \(2.35 \times 10^{-59}\) \\
                appearance & taste  & 327.685794  & \(1.15 \times 10^{-69}\) \\
                aroma      & palate & 383.575678  & \(9.83 \times 10^{-82}\) \\
                aroma      & taste  & 528.662547  & \(4.23 \times 10^{-113}\) \\
                palate     & taste  & 627.965462  & \(1.37 \times 10^{-134}\) \\ 
            \hline
        \end{tabular}%
    }
    
    \label{tab:chi_squared_results}
\end{table}

\begin{figure*}[!t]
\centering  \includegraphics[width=0.7\textwidth]{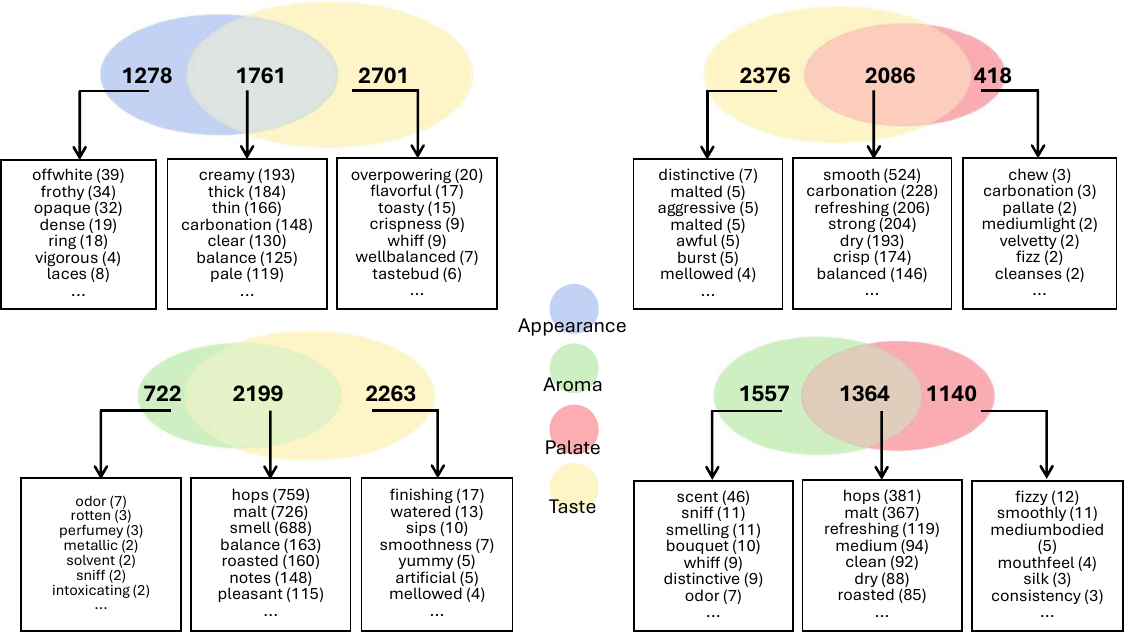}
   \caption{Statistics on lexical overlap between some aspect pairs.}
  \label{fig:sent_aspect_venn}
  \end{figure*}

Figure~\textcolor{brown}{\ref{fig:sent_aspect_venn}} illustrates the most pronounced lexical overlaps across the four aspects. Each Venn diagram shows how frequently specific descriptive terms are shared among aspects.
We observed that \texttt{taste} overlaps substantially with all other aspects. Notably, there is strong lexical overlap between \texttt{aroma–taste} and \texttt{palate–taste}, with shared terms such as ``balanced'', ``malt'', ``hop'', and ``refreshing''. This suggests that users often describe these aspects using similar linguistic patterns, reflecting their close association in the sensory experience.
In contrast, \texttt{appearance} shares fewer terms with other aspects, indicating that its vocabulary is more visually oriented and distinct.
Overall, these overlaps show that while the four aspects are conceptually distinct, their linguistic boundaries are sometimes blurred. Greater overlap indicates stronger similarity in linguistic expression, which may confound machine learning models during aspect-specific classification.

\begin{figure}[h]
    \centering
    \includegraphics[width=0.6\linewidth]{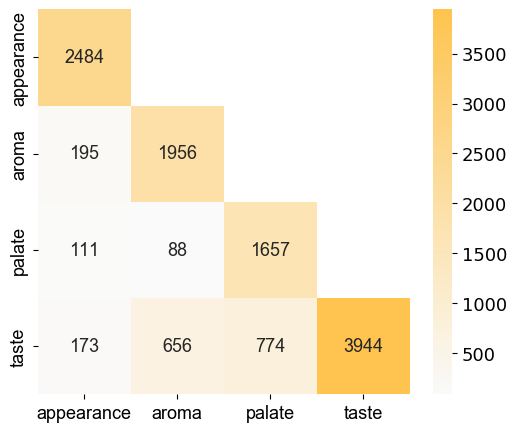}
    
    \caption{Aspect co-occurrence within a sentence}
    \label{fig:co_occurence_matrix_same_sent}
\end{figure}
\begin{figure}[h]
    \centering
    
    \includegraphics[width=0.6\linewidth]{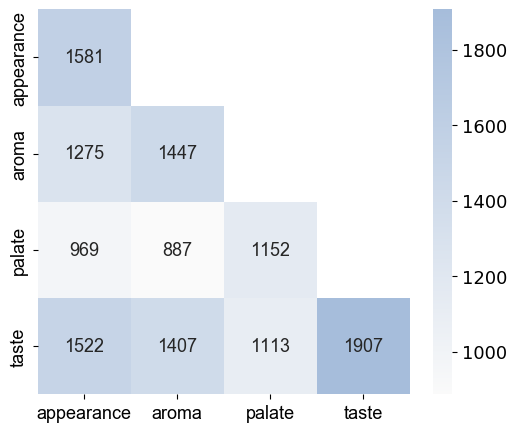}
    
    \caption{Aspect co-occurrence within a review}
    \label{fig:co_occurence_matrix_same_rev}
\end{figure}

The co-occurrence frequencies of aspect pairs in Figure~\textcolor{brown}{\ref{fig:co_occurence_matrix_same_sent}} and Figure~\textcolor{brown}{\ref{fig:co_occurence_matrix_same_rev}} reveal notable challenges for comparative opinion mining in the \dataset{} dataset.
As shown in Figure~\textcolor{brown}{\ref{fig:co_occurence_matrix_same_sent}}, at the sentence level, pairs such as \texttt{taste–aroma} and \texttt{taste–palate} frequently co-occur within the same sentence.
For instance, in the review \textit{``The taste buds were awakened nicely by the hop bitterness, yet it all finished very smoothly''}, the former clause refers only to \texttt{taste}, while the latter can pertain to both \texttt{taste} and \texttt{palate}.
Such ambiguity challenges machine learning models to determine whether an opinion targets a specific aspect or simultaneously applies to multiple ones.

At the review level (Figure~\textcolor{brown}{\ref{fig:co_occurence_matrix_same_rev}}), most reviews contain multiple aspects, indicating that users tend to evaluate beers holistically rather than focusing on a single attribute.
Among the aspect pairs, \texttt{taste–aroma} and \texttt{taste–appearance} co-occur most frequently, suggesting that reviewers often describe taste in relation to sensory cues such as smell and visual impressions. In contrast, \texttt{palate} co-appears less regularly with other aspects, implying that it is mentioned more selectively, often in contexts related to mouthfeel.
This broader co-occurrence pattern shows the need for comparative models to capture inter-aspect dependencies across sentences, as users’ opinions are distributed across multiple aspects within a review.

The chi-square values for all aspect pairs in Table~\textcolor{brown}{\ref{tab:chi_squared_results}} are exceptionally high, indicating strong interrelationships among aspects. Notably, the chi-square value between \texttt{palate} and \texttt{taste} is $627.97$, suggesting that the user evaluation of one aspect strongly predicts the other. 
For instance, reviewers who rate one aspect positively tend to rate others similarly, highlighting a potential advantage for sentiment analysis models in leveraging these correlations.
Moreover, the p-values for all pairs are extremely small, ranging from $9.20 \times 10^{-60}$ to $1.37 \times 10^{-134}$
, confirming that these relationships are statistically significant and unlikely to be due to chance.

Details on other analyses of inter-dependencies across aspects are in Appendix~\textcolor{brown}{\ref{appendix:data_analysis}}.

\section{Baseline Models \& Experimental Results}

\subsection{Baseline Models}
\begin{figure*}[!t]
  \centering
  \includegraphics[width=0.85\textwidth]{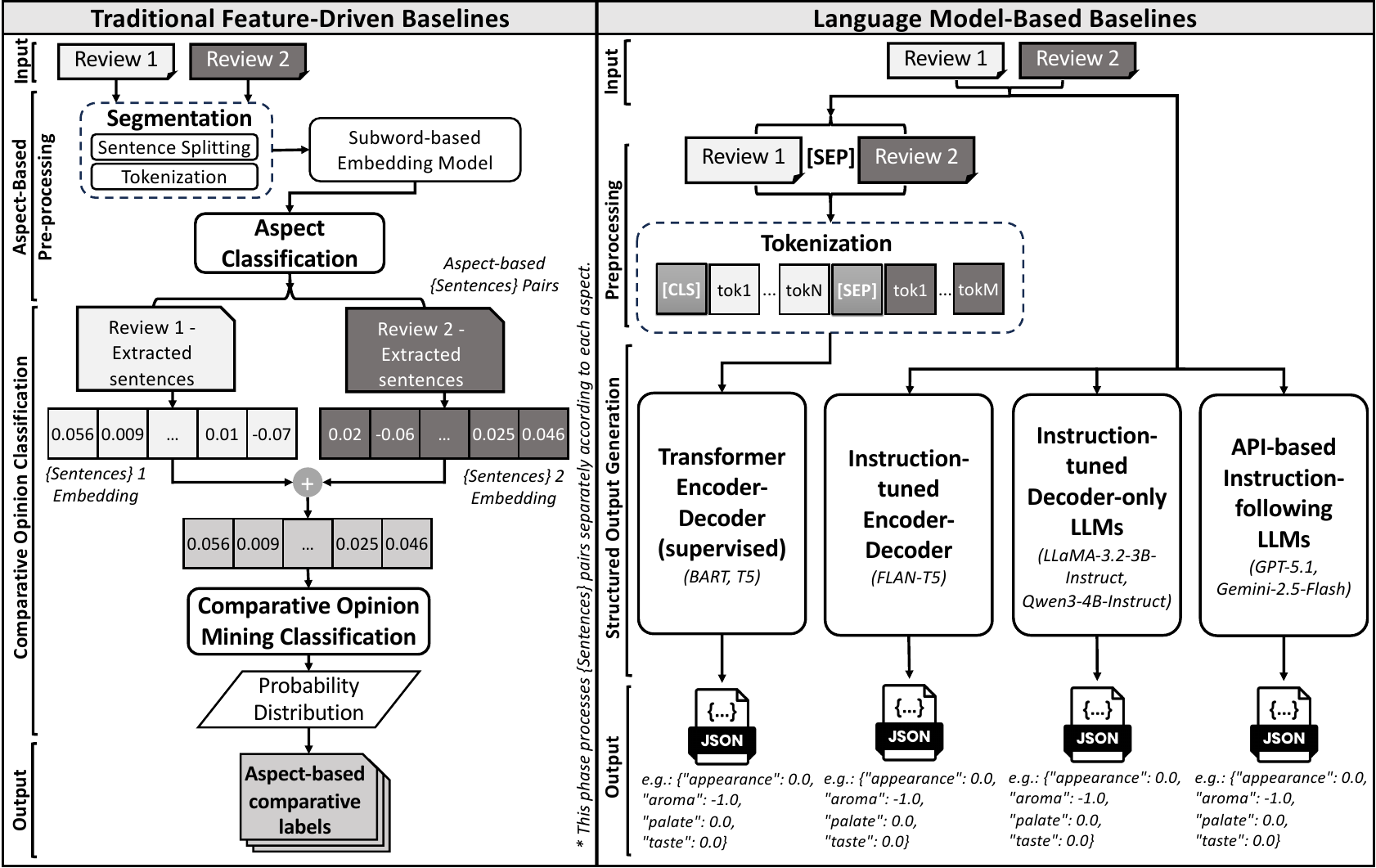}
  \caption{Baselines architecture.}
  \label{fig:model}
\end{figure*}

Figure~\textcolor{brown}{\ref{fig:model}} illustrates the benchmark models, including traditional machine learning baselines and well-known language model-based baselines commonly used in prior research.
\subsubsection{Traditional Machine Learning-Baselines}
Traditional machine learning baselines approach the task in two stages: \textit{(i)} Aspect classification and \textit{(ii)} Comparative Opinion classification. Both stages are implemented using the same traditional machine learning architecture for fair comparison.

\textbf{Pre-processing}: 
Given a dataset $\mathcal{D}=\{(R_1^u,R_2^u)\mid u\in\mathcal{U}\}$, 
each user $u$ contributes two reviews, and each review $R_i^u=[s_1^i,\dots,s_{N_i}^i]$ 
consists of $N_i$ sentences. 
We preprocessed review pairs through segmentation and tokenization and embedded using \textbf{fastText} ~\textcolor{brown}{\cite{bojanowski2017enriching}}.
For a sentence $s = [w_1, w_2, \dots, w_n]$, the embedding is computed as the mean of its  word embeddings:
\begin{equation*}
\mathbf{v}(s) = \frac{1}{n}\sum_{i=1}^{n} \mathbf{e}(w_i),
\end{equation*}
where $\mathbf{e}(w_i) \in \mathbb{R}^d$ denotes the $d$-dimensional fastText vector for word $w_i$.

\textbf{Aspect classification}: 
The first phase is considered as a multi-class, multi-label classification task. Let $\mathcal{A} = \{\textit{appearance}, \textit{aroma}, \allowbreak \textit{palate}, \textit{taste}\}$ be the predefined set of aspects. For each sentence $s$, the goal is to predict a label vector:
\[
\mathbf{y}(s) = [y_{\textit{appearance}}, y_{\textit{aroma}}, y_{\textit{palate}}, y_{\textit{taste}}] \in \{0,1\}^{|\mathcal{A}|},
\]
where $y_a = 1$ if $s$ mentions $a$, and $0$ otherwise.

\textit{Support Vector Machine (SVM):}
A separate SVM classifier $f_a$ is trained for each aspect $a$:
\[
f_a(\mathbf{v}(s)) = \text{sign}(\mathbf{w}_a^\top \mathbf{v}(s) + b_a),
\]
where $\mathbf{w}_a$ and $b_a$ are parameters for aspect $a$.

\textit{XGBoost:}
A separate XGBoost classifier $f_a$ is trained for each aspect $a$:
\begin{equation*}
    f_a(\mathbf{v}(s)) = \sigma\!\left(\sum_{k=1}^{K} f_{a,k}(\mathbf{v}(s)) \right),
\end{equation*}
where $K$ is the number of boosting rounds, $f_{a,k} \in \mathcal{F}$ denotes the $k$-th regression tree for aspect $a$, and $\sigma(\cdot)$ is the logistic function mapping the aggregated score to $[0,1]$. Each $f_{a,k}$ is trained to minimize the regularized objective:
\begin{equation*}
    \mathcal{L} = \sum_{i} \ell(y_{a,i}, \hat{y}_{a,i}) + \sum_{k} \Omega(f_{a,k}),
\end{equation*}
where $\ell$ is a differentiable loss function and $\Omega(f_{a,k})$ is the regularization term that penalizes model complexity.  
The final binary decision for each aspect is obtained by thresholding the probability:
\[
\hat{y}_a = 
\begin{cases}
1, & \text{if } f_a(\mathbf{v}(s)) \geq \tau, \\
0, & \text{otherwise}
\end{cases}
\]
where $\tau$ is the decision threshold.  

\textbf{Comparative opinion classification}: 
After aspect classification phase, the aspect-specific sentences of review pairs are extracted:
\[
s_a^1 = \text{Extract}(R_1^u, a), \quad s_a^2 = \text{Extract}(R_2^u, a),
\]
where $\text{Extract}(R, a)$ retrieves all sentences in review $R$ predicted as belonging to aspect $a$. Each review pair yields an aspect-based pair $P_a^u = (s_a^1, s_a^2)$. If either $s_a^1$ or $s_a^2$ is missing (i.e., the aspect $a$ does not appear in one of the two reviews), the pair $P_a^u$ is assigned the label \texttt{null}. This produces the structured set:
\[
\mathcal{P}^u = \{ P_a^u \mid a \in \mathcal{A} \}.
\]

The second phase aims to determine the comparative polarity between two reviews for each aspect. Given an aspect-specific pair $P_a^u = (s_a^1, s_a^2)$,  
the task is to predict a label:
\[
y_a^u \in \{\texttt{negative}, \texttt{positive}, \texttt{neutral}\},
\]
in which, for aspect $a$: \texttt{negative} means review 1 is worse than review 2, \texttt{positive} means review 1 is better than review 2, \texttt{neutral} means both reviews are equivalent.

We also used SVM and XGBoost as the classifier with the implementation principles following those outlined in the previous phase.
\begin{table*}[h]
\renewcommand{\arraystretch}{1.3}

\caption{Performance comparison between baselines. Each result represents the average of 10 independent runs. Best results are highlighted in \first{\textbf{bold}}, while second-best results are \second{\underline{underlined}}, and the third highest results are \third{marked}.}

\resizebox{\linewidth}{!}{
\begin{tabular}{llll|ccc|ccc}
\hline
\multicolumn{4}{c|}{\multirow{2}{*}{Baselines}} 
& \multicolumn{3}{c|}{\textbf{Micro-averaged}} 
& \multicolumn{3}{c}{\textbf{Macro-averaged}} \\

\multicolumn{4}{c|}{} 
& \textbf{Precision} & \textbf{Recall} & \textbf{F1} 
& \textbf{Precision} & \textbf{Recall} & \textbf{F1} \\ \hline

\multirow{6}{*}{\begin{tabular}[c]{@{}l@{}}Traditional\\ Machine\\ Learning\end{tabular}}
& \multirow{3}{*}{\begin{tabular}[c]{@{}l@{}}FastText\\ +SVM\end{tabular}}
& \multicolumn{2}{l|}{Aspect Classification}
& 90.37 & 90.37 & 90.37 & 87.68 & 88.10 & 87.84 \\ \cline{3-10}

& & \multicolumn{2}{l|}{Comparison Classification$^*$}
& 58.12 & 58.12 & 58.12 & 57.51 & 54.30 & 55.84 \\ \cline{3-10}

& & \multicolumn{2}{l|}{\textbf{Overall}}
& \second{\underline{52.57}} & 49.15 & \third{50.80} & \third{52.83} & 49.76 & \third{51.05} \\ \cline{2-10}

& \multirow{3}{*}{\begin{tabular}[c]{@{}l@{}}FastText\\ +XGBoost\end{tabular}}
& \multicolumn{2}{l|}{Aspect Classification}
& 91.24 & 91.24 & 91.24 & 88.78 & 89.11 & 88.93 \\ \cline{3-10}

& & \multicolumn{2}{l|}{Comparison Classification$^*$}
& 53.59 & 53.59 & 53.59 & 57.30 & 51.38 & 54.17 \\ \cline{3-10}

& & \multicolumn{2}{l|}{\textbf{Overall}}
& \third{52.03} & 42.28 & 46.65 & \second{\underline{53.65}} & 41.38 & 46.22 \\ \hline

\multirow{8}{*}{\begin{tabular}[c]{@{}l@{}}Language\\ Model-based\end{tabular}}
& \multirow{4}{*}{\begin{tabular}[c]{@{}l@{}}Transformer\\ Encoder-Decoder\end{tabular}}
& \multicolumn{2}{l|}{Fine-tuned BART}
& 49.70 & \first{\textbf{61.49}} & \second{\underline{54.97}} & 50.16 & \first{\textbf{62.74}} & \second{\underline{55.72}} \\ \cline{3-10}

& & \multicolumn{2}{l|}{Fine-tuned T5}
& \first{\textbf{55.88}} & \second{\underline{57.72}} & \first{\textbf{56.79}}
& \first{\textbf{56.69}} & \second{\underline{58.02}} & \first{\textbf{57.33}} \\ \cline{3-10}

& & \multicolumn{2}{l|}{Instruction-tuned FLAN-T5 (zero-shot)}
& 20.33 & 25.52 & 22.91
& 21.59 & 24.78 & 23.15 \\ \cline{3-10}

& & \multicolumn{2}{l|}{Instruction-tuned FLAN-T5 (few-shot)}
& 23.52 & 24.88 & 24.18
& 23.86 & 25.14 & 24.47 \\ \cline{2-10}

& \multirow{4}{*}{\begin{tabular}[c]{@{}l@{}}Instruction-tuned\\ Decoder-only\end{tabular}}
& \multicolumn{2}{l|}{Llama-3.2-3B-Instruct (zero-shot)}
& 24.86 & 33.90 & 28.69 & 25.62 & 34.69 & 25.61 \\ \cline{3-10}

& & \multicolumn{2}{l|}{Llama-3.2-3B-Instruct (few-shot)}
& 25.42 & 35.10 & 29.91 & 26.16 & 36.06 & 27.24 \\ \cline{3-10}

& & \multicolumn{2}{l|}{Qwen3-4B-Instruct (zero-shot)}
& 26.13 & 34.82 & 29.74 & 26.85 & 35.66 & 26.91 \\ \cline{3-10}

& & \multicolumn{2}{l|}{Qwen3-4B-Instruct (few-shot)}
& 27.43 & 36.90 & 31.81 & 28.12 & 37.54 & 29.49 \\ \cline{2-10}

& \multirow{2}{*}{\begin{tabular}[c]{@{}l@{}}API-based\\ Instruction-following\end{tabular}}
& \multicolumn{2}{l|}{ChatGPT-5.1 (zero-shot)}
& 47.28 & \third{50.33} & 48.76 & 47.61 & \third{54.33} & 47.24 \\ \cline{3-10}

& & \multicolumn{2}{l|}{Gemini-2.5-Flash (zero-shot)}
& 46.06 & 48.96 & 47.47 & 48.97 & 53.39 & 44.94 \\ \hline

\end{tabular}
}

\label{tab:baseline-results}
\end{table*}

\subsubsection{Language Model-based Baselines}
Unlike two-stage baselines, language models perform end-to-end mapping from paired reviews to structured comparisons.

\paragraph{Transformer Encoder--Decoder Models:} 
Given a pair of reviews \((R_1^u, R_2^u)\) written by user \(u\), where each review consists of a sequence of sentences, we concatenate them using a special separator token to form a single input string:
\[
\texttt{input}^{\texttt{s}} = [R_1 \oplus [\texttt{SEP}] \oplus R_2].
\]
The input is tokenized into integer token IDs $\mathbf{x} = [\texttt{tok}_1, \dots, \texttt{tok}_N, \allowbreak \texttt{[SEP]}, \texttt{tok}_1, \dots, \texttt{tok}_M]$,
where \(N\) and \(M\) denote the token lengths of \(R_1\) and \(R_2\), respectively.
We adopt \texttt{BART}~\textcolor{brown}{\cite{lewis2020bart}} and \texttt{T5}~\textcolor{brown}{\cite{raffel2020t5}}, which follow the encoder--decoder architecture and are fine-tuned on the \dataset{} training set. The encoder maps the input sequence \(\mathbf{x}\) to contextual representations \(\mathbf{h}\), while the decoder autoregressively generates a structured textual output encoding comparative sentiment decisions for predefined aspects. The generated output is post-processed into
\(\mathbf{y} = \{\text{``appearance''}: s_a, \text{``aroma''}: s_r,\)
\(\text{``palate''}: s_p, \text{``taste''}: s_t\}\),
where each sentiment score \(s_i \in \{-1.0, 0.0, 1.0, \text{None}\}\).
Model training minimizes a standard sequence-to-sequence negative log-likelihood objective:
\[
\mathcal{L} = -\sum_{t=1}^{|\mathbf{y}_{\text{seq}}|} \log P(y_{\text{seq},t} \mid y_{\text{seq},<t}, \mathbf{h}; \theta),
\]
where \(\theta\) denotes the fine-tuned parameters and \(\mathbf{y}_{\text{seq}}\) is the tokenized target sequence.

In addition, we evaluate \texttt{FLAN-T5}~\textcolor{brown}{\cite{wei2022finetuned}}, an instruction-tuned encoder--decoder model, under zero-shot and few-shot prompting, where structured outputs are generated by following instructions without fine-tuning.

\paragraph{Instruction-tuned Decoder-only Models:}
We evaluate instruction-tuned decoder-only language models, including \texttt{Llama}~\textcolor{brown}{\cite{touvron2023llama}} and \texttt{Qwen}~\textcolor{brown}{\cite{yang2024qwen}}, which generate predictions autoregressively by conditioning on the concatenated review pair and a task-specific natural language instruction. These models are evaluated under both zero-shot and few-shot prompting settings, without parameter fine-tuning.

\paragraph{API-based Instruction-following Models:}
We additionally consider API-based instruction-following models, such as \texttt{ChatGPT}~\textcolor{brown}{\cite{openai2025gpt5_system_card}} and \texttt{Gemini}~\textcolor{brown}{\cite{team2024gemini}}, which are treated as black-box systems and evaluated in a zero-shot setting using structured natural language prompts. Model outputs are parsed to obtain aspect-level comparative predictions for direct comparison.

Detailed hyper-parameter configurations of our baseline models are provided in Appendix~\textcolor{brown}{\ref{appendix:hyperparams}}.
The detailed contents of the prompts used in the zero-shot and few-shot settings are also provided in Figures~\textcolor{brown}{\ref{fig:prompt-zeroshot}} and~\textcolor{brown}{\ref{fig:prompt-fewshot}} in the Appendix.

\subsection{Experimental Results}
Evaluation is based on macro and micro averaged \texttt{Precision}, \allowbreak \texttt{Recall}, and \texttt{F1}-score, where macro treats all classes equally and micro weighs them by frequency. This evaluation offers insights into model performance under both class-balanced and class-imbalanced settings.
Table~\textcolor{brown}{\ref{tab:baseline-results}} reports the performance of traditional and language model-based baselines across evaluation metrics.
Traditional approaches using FastText with SVM or XGBoost achieve strong and stable performance on aspect classification; however, their performance drops substantially on comparative opinion classification, leading to noticeably lower overall performance.
This degradation can be attributed to their two-stage pipeline design, where errors propagate from the aspect classification stage to the comparison prediction stage, resulting in severe cascading errors.
In contrast, end-to-end language model-based baselines demonstrate superior overall performance, particularly when fine-tuned on the target task.
Notably, despite the growing popularity and strong reputation of recent instruction-tuned models, our empirical results suggest that they do not offer a performance advantage in this task setting compared to fine-tuned encoder--decoder models.
Among all evaluated baselines, fine-tuned T5 and fine-tuned BART consistently achieve the best results across both micro-averaged and macro-averaged metrics.
Detailed per-aspect results are reported in Appendix~\textcolor{brown}{\ref{appendix:detailed_results_by_aspect}}.

The overall performance is low (none exceeds $60\%$), as baselines struggle with ambiguous and overlapping aspects inherent in the \dataset{} dataset.
By focusing on same-user review pairs, our task isolates implicit comparison and reduces stylistic noise. As shown in Table~\textcolor{brown}{\ref{tab:baseline-results}}, even strong language models such as BART and T5 fail to capture subtle cross-review signals in this controlled setting. The primary challenge lies in inferring preferences from independent reviews without explicit comparative cues, with errors mainly caused by aspect ambiguity, implicit contrast, and fine-grained lexical cues.
Frequent errors, including over/under-extraction, semantic hallucination, and malformed outputs, further reflect the difficulty of modeling implicit meanings and aspect interactions (see Appendix~\textcolor{brown}{\ref{app:errors}} for detailed error analysis). These observations motivate both a dedicated benchmark for implicit comparative reasoning and the development of models with stronger semantic modeling and generalization capabilities.

\section{Conclusion}
In this paper, we present the construction of the \dataset{} dataset, annotated to capture implicit comparisons in user reviews by analyzing pairs of reviews from the same user about different products. The dataset distinguishes itself through its treatment of implicit comparisons, varied writing styles, and intricate structures. We also evaluated baselines on our dataset, revealing how well diverse algorithms perform and highlighting its robustness as a benchmark. Ultimately, testing multiple baselines establishes \dataset{} as a comprehensive resource for evaluating machine learning models and ensures its applicability across a wide range of research and practical scenarios. Our experiments show that end-to-end language model-based baselines outperform two-stage traditional machine learning baselines in generating insights, offering a valuable tool for informed decision-making. By addressing these issues, we believe future research can further enhance the effectiveness and applicability of \dataset{} and the proposed baselines in real-world comparative opinion mining tasks.

\paragraph{Limitations and Future Work}

Despite the potential of our dataset, it has several limitations. 
Firstly, our manually annotated dataset is limited in scale and domain diversity. Although the beer domain provides structured and aspect-rich content suitable for this initial study, it represents only one product category. 
Extending \dataset{} to multiple domains, such as electronics or hospitality, and incorporating a broader range of user perspectives 
would enhance the generalizability of our findings and enable a more comprehensive evaluation across diverse review styles. 
Secondly, our data analysis reveals that user reviews on different aspects often exhibit inter-dependence.
(see Section~\textcolor{brown}{\ref{subsec:aspects_interdependencies_analysis}} and Appendix~\textcolor{brown}{\ref{appendix:data_analysis}} for detailed analysis).
Specifically, users who rate one aspect positively tend to rate other aspects similarly or display a positive bias (see Section~\textcolor{brown}{\ref{subsec:aspects_interdependencies_analysis}} for detailed analysis). 
In future work, we plan to explore cross-aspect dependencies to improve model performance and enhance explainability by incorporating these interrelationships.
Finally, our baselines adopt a pipeline of separate classification modules. Future work could explore joint learning frameworks instead of modular pipelines to mitigate cascading errors. 
We believe that future research, such as developing semi-automated annotation pipelines, adopting joint learning frameworks, and establishing human performance benchmarks, can further enhance the effectiveness, interpretability, and applicability of the \dataset{} dataset as well as the baselines in real-world comparative opinion mining tasks.

\begin{acks}
We would like to sincerely thank the anonymous reviewers for their valuable feedback and insightful comments, which have greatly contributed to improving the quality of this work.
\end{acks}

\bibliographystyle{ACM-Reference-Format}
\bibliography{custom}

@article{varathan2017comparative,
  author    = {Varathan, Kasturi Dewi and Giachanou, Anastasia and Crestani, Fabio},
  title     = {Comparative Opinion Mining: A Review},
  journal   = {Journal of the Association for Information Science and Technology},
  volume    = {68},
  number    = {4},
  pages     = {811--829},
  year      = {2017},
  publisher = {Wiley Online Library}
}

@inproceedings{lewis2020bart,
  author    = {Lewis, Mike and Liu, Yinhan and Goyal, Naman and Ghazvininejad, Marjan and Mohamed, Abdelrahman and Levy, Omer and Stoyanov, Veselin and Zettlemoyer, Luke},
  title     = {BART: Denoising Sequence-to-Sequence Pre-training for Natural Language Generation, Translation, and Comprehension},
  booktitle = {Proceedings of the 58th Annual Meeting of the Association for Computational Linguistics (ACL)},
  pages     = {7871--7880},
  year      = {2020},
  publisher = {Association for Computational Linguistics}
}

@article{raffel2020t5,
  author    = {Raffel, Colin and Shazeer, Noam and Roberts, Adam and Lee, Katherine and Narang, Sharan and Matena, Michael and Zhou, Yanqi and Li, Wei and Liu, Peter J.},
  title     = {Exploring the Limits of Transfer Learning with a Unified Text-to-Text Transformer},
  journal   = {Journal of Machine Learning Research},
  volume    = {21},
  number    = {140},
  pages     = {1--67},
  year      = {2020}
}

@article{mcauley2013from,
  author        = {McAuley, Julian and Leskovec, Jure},
  title         = {From Amateurs to Connoisseurs: Modeling the Evolution of User Expertise Through Online Reviews},
  journal       = {arXiv preprint arXiv:1303.4402},
  year          = {2013},
  archivePrefix = {arXiv},
  primaryClass  = {cs.SI}
}

@inproceedings{jindal2006mining,
  author    = {Jindal, Nitin and Liu, Bing},
  title     = {Mining Comparative Sentences and Relations},
  booktitle = {Proceedings of the 21st National Conference on Artificial Intelligence (AAAI)},
  pages     = {1331--1336},
  year      = {2006},
  publisher = {AAAI Press}
}

@inproceedings{ganapathibhotla2008mining,
  author    = {Ganapathibhotla, Murthy and Liu, Bing},
  title     = {Mining Opinions in Comparative Sentences},
  booktitle = {Proceedings of the 22nd International Conference on Computational Linguistics (COLING)},
  pages     = {241--248},
  year      = {2008},
  publisher = {Association for Computational Linguistics}
}

@inproceedings{liu2021comparative,
  author    = {Liu, Zheng and Xia, Rui and Yu, Jianfei},
  title     = {Comparative Opinion Quintuple Extraction from Product Reviews},
  booktitle = {Proceedings of the 2021 Conference on Empirical Methods in Natural Language Processing (EMNLP)},
  pages     = {3955--3965},
  year      = {2021},
  publisher = {Association for Computational Linguistics}
}

@inproceedings{hayate2022comparative,
  author    = {Iso, Hayate and Wang, Xiaolan and Angelidis, Stefanos and Suhara, Yoshihiko},
  title     = {Comparative Opinion Summarization via Collaborative Decoding},
  booktitle = {Findings of the Association for Computational Linguistics (ACL)},
  year      = {2022},
  publisher = {Association for Computational Linguistics}
}

@article{can2025toward,
  author    = {Can, Duy-Cat and Nguyen, Khanh-Vinh and Hoang, Hung-Manh and Vu, Duc-Loc and Tran, Mai-Vu and Le, Hoang-Quynh},
  title     = {Toward Effective Comparative Opinion Mining: A Novel Vietnamese Product Review Corpus and Benchmark Approach},
  journal   = {IEEE Access},
  volume    = {13},
  pages     = {82111--82128},
  year      = {2025},
  publisher = {IEEE}
}

@mastersthesis{steinert2017collaborative,
  author    = {Steinert, Kai},
  title     = {Collaborative Web-Based Short Text Annotation with Online Label Suggestion},
  school    = {TU Darmstadt},
  type      = {Master's Thesis},
  year      = {2017}
}

@article{bojanowski2017enriching,
  author    = {Bojanowski, Piotr and Grave, Edouard and Joulin, Armand and Mikolov, Tomas},
  title     = {Enriching Word Vectors with Subword Information},
  journal   = {Transactions of the Association for Computational Linguistics},
  volume    = {5},
  pages     = {135--146},
  year      = {2017}
}

@article{touvron2023llama,
  title={LLaMA: Open and Efficient Foundation Language Models},
  author={Touvron, Hugo and others},
  journal={arXiv preprint arXiv:2302.13971},
  year={2023}
}

@article{yang2024qwen,
  title={Qwen Technical Report},
  author={Yang, Jinrui and others},
  journal={arXiv preprint arXiv:2309.16609},
  year={2024}
}

@misc{openai2025gpt5_system_card,
  title        = {GPT-5 System Card},
  author       = {{OpenAI}},
  howpublished = {Technical Report, OpenAI},
  year         = {2025},
  url          = {https://cdn.openai.com/gpt-5-system-card.pdf}
}

@article{team2024gemini,
  title={Gemini: A Family of Highly Capable Multimodal Models},
  author={{Google DeepMind}},
  journal={arXiv preprint arXiv:2312.11805},
  year={2024}
}

@article{wei2022finetuned,
  title={Finetuned Language Models Are Zero-Shot Learners},
  author={Wei, Jason and others},
  journal={arXiv preprint arXiv:2109.01652},
  year={2022}
}

\appendix




\section{Detailed Results by Aspect}
\label{appendix:detailed_results_by_aspect}

\begin{table*}[h]
\centering
\footnotesize
\renewcommand{\arraystretch}{1.3}

\caption{Baselines performance comparison across four aspects (\%).}

\begin{tabular}{p{3.0cm}|lcccccc}
\hline
\textbf{Model} & \textbf{Aspect}
& \makecell{\textbf{Micro-Avg}\\\textbf{Precision}}
& \makecell{\textbf{Micro-Avg}\\\textbf{Recall}}
& \makecell{\textbf{Micro-Avg}\\\textbf{F1}}
& \makecell{\textbf{Macro-Avg}\\\textbf{Precision}}
& \makecell{\textbf{Macro-Avg}\\\textbf{Recall}}
& \makecell{\textbf{Macro-Avg}\\\textbf{F1}} \\
\hline

\multirow{4}{*}{FastText + SVM}
& Appearance & 54.26 & 53.65 & 53.95 & 57.42 & 53.49 & 54.49 \\
& Aroma      & 57.60 & 52.16 & 54.76 & 57.92 & 55.05 & 55.99 \\
& Palate    & 55.25 & 43.75 & 48.85 & 55.37 & 45.15 & 48.78 \\
& Taste     & 43.66 & 42.47 & 43.05 & 45.00 & 43.07 & 44.02 \\
\hline

\multirow{4}{*}{FastText + XGBoost}
& Appearance & 50.50 & 49.93 & 50.18 & 58.33 & 47.78 & 48.73 \\
& Aroma      & 54.35 & 49.21 & 51.65 & 55.92 & 45.82 & 49.88 \\
& Palate    & 55.25 & 43.75 & 48.45 & 57.13 & 42.22 & 47.69 \\
& Taste     & 45.30 & 44.07 & 44.68 & 45.28 & 45.45 & 45.36 \\
\hline

\multirow{4}{*}{Fine-tuned BART}
& Appearance & 57.53 & 62.45 & 59.89 & 58.11 & 64.66 & 60.91 \\
& Aroma      & 50.96 & 67.09 & 57.92 & 56.22 & 66.63 & 60.18 \\
& Palate    & 26.85 & 48.07 & 34.46 & 27.90 & 49.08 & 35.06 \\
& Taste     & 61.92 & 63.73 & 62.81 & 61.93 & 64.75 & 63.16 \\
\hline

\multirow{4}{*}{Fine-tuned T5}
& Appearance & 63.54 & 65.43 & 64.47 & 64.78 & 66.02 & 65.33 \\
& Aroma      & 60.17 & 59.92 & 60.04 & 61.38 & 59.21 & 60.08 \\
& Palate    & 44.79 & 47.51 & 46.11 & 45.15 & 46.45 & 45.57 \\
& Taste     & 53.32 & 55.73 & 54.50 & 53.75 & 56.53 & 55.09 \\
\hline

\multirow{4}{*}{\makecell[l]{Instruction-tuned\\FLAN-T5 (zero-shot)}}
& Appearance & 23.41 & 24.02 & 23.71 & 23.88 & 24.34 & 24.10 \\
& Aroma      & 22.96 & 23.43 & 23.19 & 23.32 & 23.75 & 23.53 \\
& Palate     & 21.31 & 21.87 & 21.58 & 21.66 & 22.02 & 21.83 \\
& Taste      & 22.22 & 22.75 & 22.48 & 22.61 & 23.01 & 22.80 \\
\hline

\multirow{4}{*}{\makecell[l]{Instruction-tuned\\FLAN-T5 (few-shot)}}
& Appearance & 24.86 & 25.42 & 25.13 & 25.19 & 25.73 & 25.45 \\
& Aroma      & 24.31 & 24.84 & 24.57 & 24.68 & 25.09 & 24.88 \\
& Palate     & 22.66 & 23.21 & 22.93 & 22.89 & 23.35 & 23.07 \\
& Taste      & 23.94 & 24.53 & 24.23 & 24.18 & 24.71 & 24.42 \\
\hline

\multirow{4}{*}{\makecell[l]{LLaMA-3.2-3B-Instruct\\(zero-shot)}}
& Appearance & 28.29 & 36.80 & 31.99 & 29.02 & 38.06 & 27.27 \\
& Aroma      & 22.22 & 31.22 & 25.96 & 23.45 & 32.21 & 22.36 \\
& Palate     & 16.31 & 33.70 & 21.98 & 16.90 & 32.69 & 20.85 \\
& Taste      & 32.23 & 33.60 & 32.90 & 36.45 & 34.59 & 29.52 \\
\hline

\multirow{4}{*}{\makecell[l]{LLaMA-3.2-3B-Instruct\\(few-shot)}}
& Appearance & 27.88 & 36.72 & 31.60 & 28.45 & 37.18 & 29.42 \\
& Aroma      & 26.41 & 34.96 & 30.01 & 26.89 & 35.48 & 27.73 \\
& Palate     & 23.62 & 32.84 & 27.43 & 24.01 & 33.29 & 25.01 \\
& Taste      & 25.98 & 35.21 & 29.42 & 26.31 & 36.31 & 26.79 \\
\hline

\multirow{4}{*}{\makecell[l]{Qwen3-4B-Instruct\\(zero-shot)}}
& Appearance & 27.42 & 35.68 & 31.00 & 27.95 & 36.11 & 27.83 \\
& Aroma      & 26.03 & 34.21 & 29.43 & 26.51 & 34.75 & 26.62 \\
& Palate     & 24.19 & 32.75 & 27.77 & 24.58 & 33.18 & 25.04 \\
& Taste      & 26.12 & 35.14 & 29.86 & 26.37 & 35.63 & 27.14 \\
\hline

\multirow{4}{*}{\makecell[l]{Qwen3-4B-Instruct\\(few-shot)}}
& Appearance & 29.41 & 37.92 & 33.12 & 30.02 & 38.46 & 30.61 \\
& Aroma      & 28.03 & 36.31 & 31.54 & 28.49 & 36.85 & 28.97 \\
& Palate     & 26.11 & 34.72 & 29.31 & 26.54 & 35.18 & 27.04 \\
& Taste      & 28.17 & 37.03 & 31.91 & 28.43 & 37.65 & 29.34 \\
\hline

\multirow{4}{*}{ChatGPT-5.1 (zero-shot)}
& Appearance & 55.76 & 56.18 & 55.97 & 55.16 & 59.66 & 53.18 \\
& Aroma      & 52.36 & 47.23 & 49.66 & 56.01 & 52.61 & 49.50 \\
& Palate     & 35.51 & 48.33 & 40.94 & 35.24 & 51.85 & 40.40 \\
& Taste      & 46.10 & 49.06 & 47.53 & 46.01 & 52.44 & 44.95 \\
\hline

\multirow{4}{*}{\makecell[l]{Gemini-2.5-Flash\\(zero-shot)}}
& Appearance & 52.61 & 52.42 & 52.51 & 54.79 & 56.48 & 48.24 \\
& Aroma      & 53.88 & 52.74 & 53.30 & 57.73 & 57.65 & 53.04 \\
& Palate     & 33.19 & 41.44 & 36.86 & 31.65 & 46.03 & 33.96 \\
& Taste      & 44.42 & 47.73 & 46.02 & 47.16 & 51.52 & 42.05 \\
\hline

\multicolumn{8}{r}{\textit{Macro-Avg: macro-averaged, Micro-Avg: micro-averaged.}}\\
\end{tabular}

\label{tab:aspect-results}
\end{table*}

Among language model-based baselines, end-to-end fine-tuned models consistently outperform both traditional methods and instruction-tuned large language models (Table~\textcolor{brown}{\ref{tab:aspect-results}}). Specifically, \texttt{fine-tuned BART} achieves \texttt{Macro-F1} scores of $60.91$, $60.18$, $35.06$, and $63.16$ across the four aspects, while \texttt{fine-tuned T5} further improves performance on most aspects, reaching $65.33$ on \texttt{Appearance}, $60.08$ on \texttt{Aroma}, $45.57$ on \texttt{Palate}, and $55.09$ on \texttt{Taste}.
In contrast, instruction-tuned models such as \texttt{FLAN-T5}, \texttt{LLaMA-3.2}, and \texttt{Qwen3}, evaluated under zero-shot and few-shot settings, yield substantially lower \texttt{Macro-F1} scores, typically remaining below $35\%$ across all aspects. This highlights the limitation of relying solely on instruction-following capabilities without task-specific fine-tuning.

\section{Error Analysis}
\label{subsec:error-analysis}
Although our baseline models achieve reasonable performance, they consistently struggle with various challenging cases, underscoring the complexity of the dataset. Table~\textcolor{brown}{\ref{tab:error-analysis}} summarizes some common error types and representative examples from \dataset{} dataset.

A common issue with traditional machine learning-baselines is the over-extraction of aspect-related sentences, where models mistakenly include text that is too generic or only weakly relevant (e.g., ``\textit{This beer is very easy to drink}''), highlighting the challenge of detecting aspect-specific content in subjective language.
Conversely, under-extraction often arises when aspect-relevant information is conveyed implicitly, as in \textit{``goes down smooth and tasty}'', which is not detected due to the absence of explicit aspect markers.

Language model-based baselines alleviate many surface-level issues through contextual embeddings, but they often rely excessively on learned patterns rather than actual content, leading to inaccurate predictions.
One notable issue is JSON structure corruption, where a model produces incomplete or malformed outputs (e.g., missing aspect fields).
Another frequent error is semantic hallucination, in which the model predicts aspects that are never mentioned in the review.

Some errors are shared by both traditional machine learning- and language model-based baselines. Aspect ambiguity is a major challenge: certain terms (e.g., ``\textit{hops}'') may refer to multiple aspects such as \texttt{taste} or \texttt{aroma}, leading to confusion. Co-occurring aspects in the same sentence can also cause sentiment interference (e.g., \texttt{appearance} influencing \texttt{taste}). These issues highlight persistent challenges in comparative opinion mining, handling implicit or ambiguous aspect cues and maintaining structural consistency in outputs.



\label{app:errors}

\begin{table*}[h]
\centering
\small
\renewcommand{\arraystretch}{1.4}
\resizebox{\textwidth}{!}{%
\begin{tabular}{l|c|c|l}
\hline
\textbf{Error type} & \textbf{\begin{tabular}[c]{@{}c@{}}Traditional Machine\\ Learning-$^*$\end{tabular}} & \textbf{\begin{tabular}[c]{@{}c@{}}Language Model-\\ based$^+$\end{tabular}} & \textbf{Representative examples} \\ \hline
\begin{tabular}[c]{@{}l@{}}Over-extraction \\ of aspect-related \\ sentence\end{tabular} & \checkmark &  & \begin{tabular}[c]{@{}l@{}}``\textit{This beer is very easy to drink}''\\ The sentence is general and not truly about a specific aspect.\end{tabular} \\ \hline
\begin{tabular}[c]{@{}l@{}}Under-extraction\\ of aspect-related\\ sentence\end{tabular} & \checkmark &  & \begin{tabular}[c]{@{}l@{}}``\textit{I like to get it from time to time when I want something with style that} \\ \textit{\textbf{goes down smooth and tasty}}'' \\ Missed implicit aspect.\end{tabular} \\ \hline
\begin{tabular}[c]{@{}l@{}}JSON structure\\ corruption\end{tabular} &  & \checkmark & \begin{tabular}[c]{@{}l@{}}\textit{Output: \{``appearance'': 1, ``taste'': 2\}}\\ The model outputs an incomplete JSON object, missing aroma and palate.\end{tabular} \\ \hline
\begin{tabular}[c]{@{}l@{}}Semantic\\ hallucination\end{tabular} &  & \checkmark & \begin{tabular}[c]{@{}l@{}}\textit{Review 1: ``Pours a bright amber \ldots palate is very balanced.'' }\\ \textit{Review 2: ``Dark brown appearance \ldots Taste is outstanding''} \\ \textbf{No aroma mentioned}, yet predicted.\end{tabular} \\ \hline
Aspect ambiguity & \checkmark & \checkmark & \begin{tabular}[c]{@{}l@{}}``\textit{\textbf{Hops} is in the background but noticeable}'' \\ Model confused between aroma and taste.\end{tabular} \\ \hline
\begin{tabular}[c]{@{}l@{}}Cross-aspect\\ sentiment \\ interference\end{tabular} & \checkmark & \checkmark & \begin{tabular}[c]{@{}l@{}}``\textit{The \textbf{lace} is amazing and makes this beer almost as good to watch as to} \\ \textit{Taste (well, not quite but you get the idea)}'' \\ The model mistakenly predicts a highly positive sentiment for taste.\end{tabular} \\ \hline
\multicolumn{4}{r}{*: Traditional Machine Learning-baselines, +: Language Model-based baselines.}
\end{tabular}%
}

\caption{Examples of representative error types observed across different baseline models.}
\small{(\checkmark) indicates that the corresponding model type exhibits the listed error.}

\label{tab:error-analysis}
\end{table*}

\section{More Data Detailed Statistics and Analysis}
\label{appendix:data_analysis}

\begin{figure}[h]
       \centering
        \includegraphics[width=0.8\linewidth]{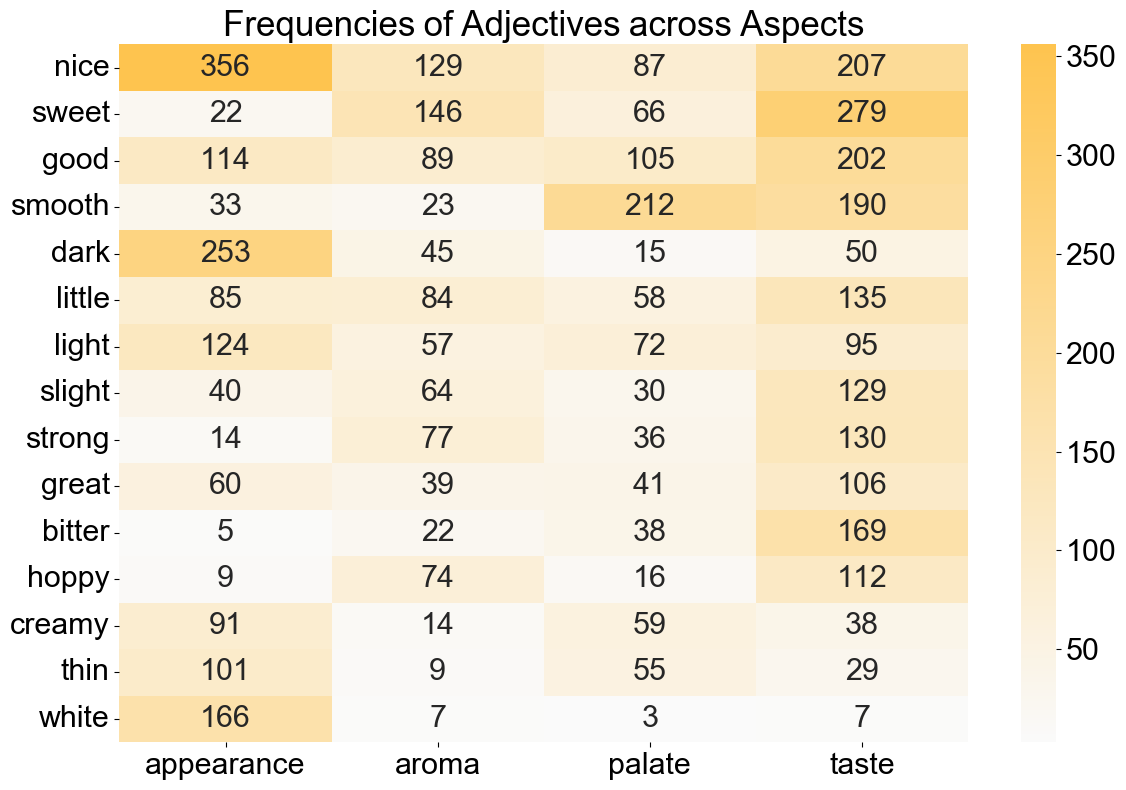}  
    \caption{Statistics on occurrence counts of adjectives appeared across aspects of the \dataset{} dataset.}
    \label{fig:adj_heatmap}
    \vspace{-2em}
\end{figure}
\begin{figure}[h]
    \centering
     \includegraphics[width=\linewidth]{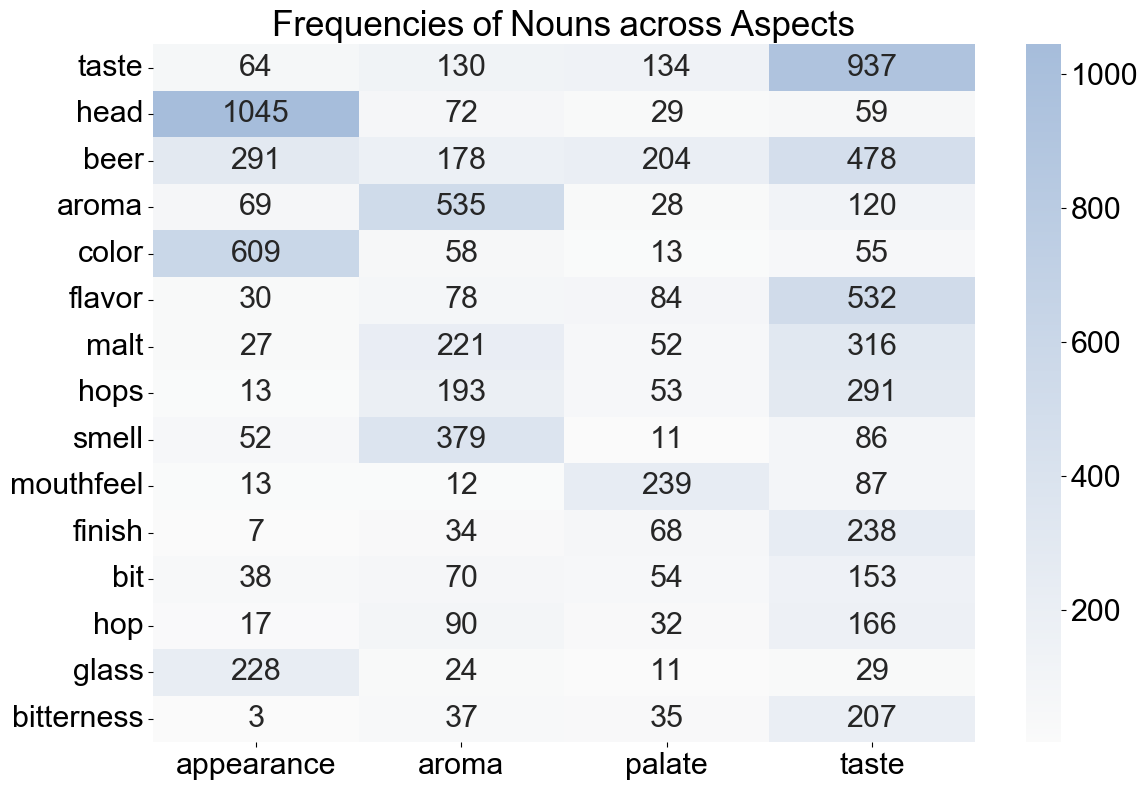} 
    \caption{Statistics on occurrence counts of nouns appeared across aspects of the \dataset{} dataset.}
    \label{fig:noun_heatmap}
\end{figure}

Figures~\textcolor{brown}{\ref{fig:adj_heatmap}} and~\textcolor{brown}{\ref{fig:noun_heatmap}} illustrate the top-frequency adjectives and nouns appearing for each aspect. The adjective heatmap reveals considerable overlaps, as an adjective can describe multiple aspects. The noun heatmap, however, simplifies the classifier's task by associating distinct sets of nouns with specific aspects, making aspect identification in review sentences less ambiguous.

\section{Training Environment and Hyper-parameter Configurations}
\label{appendix:hyperparams}
This appendix details the computational environment, software stack, and hyper-parameter configurations used for training and evaluating all baseline models.

\subsection{Hardware and Software Environment}

Experiments are conducted on the Kaggle environment with the following hardware specifications:
\begin{itemize}[noitemsep]
    \item CPU: 2x vCPU Intel Xeon (2.20GHz)
    \item GPU: NVIDIA Tesla P100 (16GB VRAM)
    \item RAM: 30 GB
    \item Storage: 58 GB
\end{itemize}

The software environment is standardized as follows:
\begin{itemize}[noitemsep]
    \item Python: 3.11.13
    \item torch: 2.6.0+cu124
    \item nltk: 3.9.1
    \item transformers: 4.52.4
    \item datasets: 3.6.0
    \item scikit-learn: 1.2.2
    \item xgboost: 2.0.3
    \item fasttext-wheel: 0.9.2
    \item numpy: 1.26.4
    \item pandas: 2.2.3
    \item scipy: 1.15.3
    \item tqdm: 4.67.1
    \item matplotlib: 3.7.2
    \item seaborn: 0.12.2
    \item imblearn: 0.0
    \item scikit-learn: 1.6.1
    \item huggingface-hub: 0.33.1
\end{itemize}

\subsection{Training Settings}

The following hyper-parameters were used as default settings for all models unless otherwise specified:

\begin{itemize}[noitemsep]
    \item Optimizer: AdamW
    \item Learning rate: $2 \times 10^{-4}$
    \item Weight decay: 0.01
    \item Batch size: 4
    \item Gradient accumulation steps: 4
    \item Number of epochs: 15
    \item Learning rate scheduler: Linear decay with warmup
    \item Warmup steps: 500
    \item Early stopping: 5 epochs
    \item Random seed: 42
\end{itemize}

\subsection{Model-Specific Hyper-parameters}

\subsubsection{FastText+SVM Baseline}
\paragraph{Aspect-Based Pre-processing.}
\begin{itemize}[noitemsep]
    \item Pretrained checkpoint: 'bert-base-uncased'
    \item Maximum sequence length: 128
    \item Hidden size: 768
    \item Number of transformer layers: 12
    \item Number of attention heads: 12
\end{itemize}

\paragraph{Comparative Opinion Classification.}

We employed FastText embeddings and XGBoost for comparative opinion classification:  
\begin{itemize}[noitemsep]
    \item Encoder: pretrained FastText checkpoint  
    \begin{itemize}[noitemsep]
        \item Downloaded from: \url{https://dl.fbaipublicfiles.com/fasttext/vectors-crawl/cc.en.300.bin.gz}  
        \item Embedding dimension: 300  
        \item Window size: 5  
        \item Minimum count: 5  
        \item Epochs: 5  
    \end{itemize}
\end{itemize}

The best hyper-parameters (determined via Grid Search) for each aspect are summarized in Table~\textcolor{brown}{\ref{tab:comp-svm-results}}.

\begin{table}[H]
\centering
\small
\renewcommand{\arraystretch}{1.3}

\caption{Best hyper-parameters of SVM for each aspect.}

\begin{tabular}{l|l}
\hline
\textbf{Aspect} & \textbf{SVM} \\ \hline
Appearance & \begin{tabular}[t]{@{}l@{}}kernel=rbf \\ C=10 \\ $\gamma$=scale\end{tabular} \\ \hline
Aroma      & \begin{tabular}[t]{@{}l@{}}kernel=rbf \\ C=1 \\ $\gamma$=scale\end{tabular}  \\ \hline
Palate     & \begin{tabular}[t]{@{}l@{}}kernel=linear \\ C=10 \\ $\gamma$=scale\end{tabular} \\ \hline
Taste      & \begin{tabular}[t]{@{}l@{}}kernel=linear \\ C=10 \\ $\gamma$=scale\end{tabular} \\ \hline
\end{tabular}

\label{tab:comp-svm-results}
\end{table}

\subsubsection{FastText+XGBoost Baseline}
\paragraph{Aspect-Based Pre-processing.}
\begin{itemize}[noitemsep]
    \item Pretrained checkpoint: 'bert-base-uncased'
    \item Maximum sequence length: 128
    \item Hidden size: 768
    \item Number of transformer layers: 12
    \item Number of attention heads: 12
\end{itemize}

\paragraph{Comparative Opinion Classification.}
We employed FastText embeddings and XGBoost for comparative opinion classification:  
\begin{itemize}[noitemsep]
    \item Encoder: pretrained FastText checkpoint  
    \begin{itemize}[noitemsep]
        \item Downloaded from: \url{https://dl.fbaipublicfiles.com/fasttext/vectors-crawl/cc.en.300.bin.gz}  
        \item Embedding dimension: 300  
        \item Window size: 5  
        \item Minimum count: 5  
        \item Epochs: 5  
    \end{itemize}
\end{itemize}

For the classifier, we adopted XGBoost. The hyper-parameters were tuned separately for each aspect using Grid Search.  
The best configurations for each aspect are summarized in Table~\textcolor{brown}{\ref{tab:xgb-results}}.

\begin{table}[H]
\centering
\small
\renewcommand{\arraystretch}{1.3}

\caption{Best XGBoost hyperparameters for each aspect.}

\begin{tabular}{l|l}
\hline
\textbf{Aspect} & \textbf{XGBoost Params} \\ \hline
Appearance & \begin{tabular}[t]{@{}l@{}}max\_depth=5 \\ eta=0.1 \\ subsample=1.0 \\ colsample\_bytree=0.8 \\ min\_child\_weight=3 \\ \end{tabular} \\ \hline
Aroma & \begin{tabular}[t]{@{}l@{}}max\_depth=3 \\ eta=0.1 \\ subsample=0.8 \\ colsample\_bytree=0.8 \\ min\_child\_weight=1 \\ \end{tabular} \\ \hline
Palate & \begin{tabular}[t]{@{}l@{}}max\_depth=7 \\ eta=0.1 \\ subsample=0.8 \\ colsample\_bytree=1.0 \\ min\_child\_weight=5 \\ \end{tabular} \\ \hline
Taste & \begin{tabular}[t]{@{}l@{}}max\_depth=3 \\ eta=0.1 \\ subsample=0.8 \\ colsample\_bytree=0.8 \\ min\_child\_weight=1 \\ \end{tabular} \\ \hline
\end{tabular}

\label{tab:xgb-results}
\end{table}

\subsubsection{Fine-tuned T5}
\begin{itemize}[noitemsep]
    \item Pretrained checkpoint: 'google-t5/t5-large'
    \item Maximum sequence length: 880
    \item Hidden size: 1024
    \item Number of transformer layers: 24
    \item Number of attention heads: 16
\end{itemize}

\subsubsection{Fine-tuned BART}
\begin{itemize}[noitemsep]
    \item Pretrained checkpoint: 'facebook/bart-large'
    \item Maximum sequence length: 880
    \item Hidden size: 1024
    \item Number of transformer layers: 12 encoder + 12 decoder
    \item Number of attention heads: 16
\end{itemize}

\subsubsection{FLAN-T5}
\begin{itemize}[noitemsep]
    \item Pretrained checkpoint: `google/flan-t5-large`
    \item Maximum sequence length: 880
    \item Hidden size: 1024
    \item Number of transformer layers: 24
    \item Number of attention heads: 16
\end{itemize}

\subsubsection{LLaMA-3.2-3B-Instruct}
\begin{itemize}[noitemsep]
    \item Pretrained checkpoint: `meta-llama/LLaMA-3.2-3B-Instruct`
    \item Maximum sequence length: 128000
    \item Hidden size: 3,072
    \item Number of transformer layers: 28
    \item Number of attention heads: 24
\end{itemize}

\subsubsection{Qwen3-4B-Instruct}
\begin{itemize}[noitemsep]
    \item Pretrained checkpoint: `Qwen/Qwen3-4B-Instruct`
    \item Maximum sequence length: 32,768
    \item Hidden size: 4,096
    \item Number of transformer layers: 36
    \item Number of attention heads: 40
\end{itemize}

\subsubsection{ChatGPT-5.1}
\begin{itemize}[noitemsep]
    \item Model: ChatGPT-5.1 (OpenAI API)
    \item Maximum sequence length: Not publicly available
    \item Hidden size: Not publicly available
    \item Number of transformer layers: Not publicly available
    \item Number of attention heads: Not publicly available
\end{itemize}

\subsubsection{Gemini-2.5-Flash}
\begin{itemize}[noitemsep]
    \item Model: Gemini-2.5-Flash (Google API)
    \item Maximum sequence length: Not publicly available
    \item Hidden size: Not publicly available
    \item Number of transformer layers: Not publicly available
    \item Number of attention heads: Not publicly available
\end{itemize}

\subsection{Training Time}

Across all four aspects, traditional machine learning-baselines required about 6 hours, mainly for Aspect-Based Pre-processing, while Comparative Opinion Classification step took only a few minutes.  
By comparison, both fine-tuned T5 and fine-tuned BART required approximately 7 hours to complete end-to-end training.

\begin{figure*}[h]
    \centering
    \includegraphics[width=\textwidth]{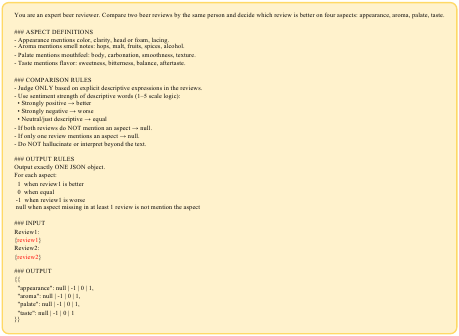}
    \caption{Zero-shot prompt used for instruction-following models.}
    \label{fig:prompt-zeroshot}
\end{figure*}

\begin{figure*}[h]
    \centering
    \includegraphics[width=0.95\textwidth]{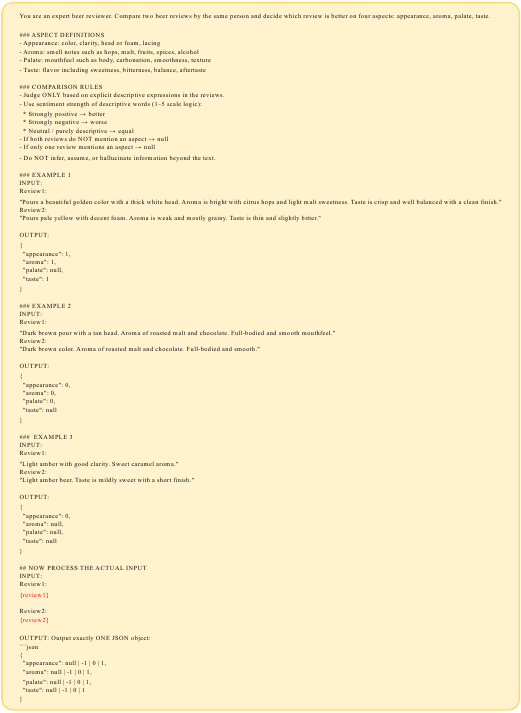}
    \caption{Few-shot prompt used for instruction-following models.}
    \label{fig:prompt-fewshot}
\end{figure*}

\end{document}